%% file: main.tex
\documentclass[runningheads]{llncs}
\usepackage{graphicx}
\usepackage{amsmath,amssymb} % define this before the line numbering.
\usepackage{cite,url}
\usepackage{color}
\usepackage{enumerate}
\usepackage[width=122mm,left=12mm,paperwidth=146mm,height=193mm,top=12mm,paperheight=217mm]{geometry}
\begin{document}
% \renewcommand\thelinenumber{\color[rgb]{0.2,0.5,0.8}\normalfont\sffamily\scriptsize\arabic{linenumber}\color[rgb]{0,0,0}}
% \renewcommand\makeLineNumber {\hss\thelinenumber\ \hspace{6mm} \rlap{\hskip\textwidth\ \hspace{6.5mm}\thelinenumber}}
% \linenumbers
\pagestyle{headings}
\mainmatter
\def\ECCV18SubNumber{***}  % Insert your submission number here

\newcommand\gp{\mathrm{GAP}}
\newcommand\gap[1]{\mathrm{GAP@}{#1}}

\title{Label Denoising with Large Ensembles of Heterogeneous Neural Networks} % Replace with your title

\titlerunning{Label Denoising with Large Ensembles of Heterogeneous Neural Networks}

\authorrunning{Ostyakov, Logacheva, Suvorov, Aliev, Sterkin, Khomenko, Nikolenko}

\author{Pavel Ostyakov$^1$ \and Elizaveta Logacheva$^1$
	\and Roman Suvorov$^1$ \and Vladimir Aliev$^1$
	\and Gleb Sterkin$^1$
	Oleg Khomenko$^1$
	\and Sergey I. Nikolenko$^{1,2,3}$\\
}
\institute{
	$^1$Samsung AI Center, Moscow, Russia\\
\texttt{\small \{p.ostyakov, r.suvorov, e.logacheva, o.khomenko\}@samsung.com} \\
$^2$Steklov Mathematical Institute at St. Petersburg\\
$^3$Neuromation OU, Tallinn, Estonia\\
\texttt{\small sergey@logic.pdmi.ras.ru}
}

\maketitle

\begin{abstract}
Despite recent advances in computer vision based on various convolutional architectures, video understanding remains an important challenge. In this work, we present and discuss a top solution for the large-scale video classification (labeling) problem introduced as a Kaggle competition based on the YouTube-8M dataset. We show and compare different approaches to preprocessing, data augmentation, model architectures, and model combination. Our final model is based on a large ensemble of video- and frame-level models but fits into rather limiting hardware constraints. We apply an approach based on knowledge distillation to deal with noisy labels in the original dataset and the recently developed mixup technique to improve the basic models.

\end{abstract}

\section{Introduction}

Video understanding and learning high-quality latent representations for videos is an important problem which largely remains open and needs more advances in computer vision, including video processing and scene understanding, audio processing and speech recognition, and natural language processing. One of the first steps towards true video understanding is video classification/labeling, where the problem is to apply labels from a predefined set to a video. While, e.g., text classifiers achieve very high results despite the fact that general language understanding remains a hard problem, for video data even classification is relatively underexplored.

In this work, we concentrate on video classification task presented in the second YouTube-8M Challenge, where the problem is to automatically annotate YouTube videos with a large number of predefined labels based on video-level and frame-level features. While we have introduced several novel modifications to existing deep learning models and performed a large-scale experimental comparison of convolutional, recurrent, and attention-based architectures, we believe that our main contribution lies in dealing with noisy labels. One of the main challenges in the YouTube-8M dataset and other large-scale datasets is that the labels, while relatively high-quality, are not completely reliable. In fact, according to the authors of the dataset, precision and recall of YouTube-8M labels is 78.8\% and 14.5\% respectively compared to the gold standard of human raters~\cite{DBLP:journals/corr/Abu-El-HaijaKLN16}. In this work, we deal with this problem with the recently developed knowledge distillation approach which we base on a large ensemble of a wide variety of video- and frame-level models. The ensemble, however, is used only for preparing soft labels, and the final model trained on these soft labels is relatively simple and fits into rather limiting hardware requirements imposed in the challenge. Our final resulting model achieved very high results in video classification and placed 4th in the second YouTube-8M Challenge.

The paper is organized as follows. In Section~\ref{sec:data}, we describe the dataset, introduce the primary evaluation metric that we use throughout the paper, and discuss different data representation methods that we compared for frame-level data. Section~\ref{sec:models} describes the main neural architectures used in this work and presents the ensemble-based distillation that we used to alleviate the problem of noisy labels. Section~\ref{sec:eval} presents an extensive evaluation study of the proposed models and a detailed error analysis, and Section~\ref{sec:concl} concludes the paper.

\section{Data and Evaluation}\label{sec:data}

\subsubsection*{Dataset.}
The YouTube-8M dataset contains 5.6M videos encoded as a hidden representation produced by Deep CNN pretrained on the ImageNet dataset~\cite{imagenet_cvpr09} for both audio spectrogram and video frames taken at rate of 1Hz (once per second). The dataset also contains aggregated \emph{video-level features} extracted as averaged frame-level features. Each video was automatically annotated with $3862$ classes by the YouTube video annotation system. To ensure high quality of the videos in the dataset, videos were preselected to meet the following requirements: each video must be public, have at least $1000$ views, be between $120$ and $500$ seconds long, and also be associated with at least one entity from the target vocabulary; adult and sensitive content has been removed based on automated classifiers. The dataset is an improved version of the initially released YouTube-8M~\cite{DBLP:journals/corr/Abu-El-HaijaKLN16}, with some noisy and rare labels filtered out, but the labeling still remains rather noisy.

\subsubsection*{Evaluation.}
The primary evaluation metric for our experiments and the competition itself is the \emph{global average precision} (GAP), defined as $\gap{n} = \sum_{i=1}^N p(i) r(i)$, where $N$ is the total number of final predictions, $p(i)$ is the precision score at rank $i$, and $r(i)$ denotes the relevance of prediction $i$: it’s $1$ if the $i$-th prediction is correct, and $0$ otherwise. For every video top $n$ predicted labels are scored, so $N=nV$, where $V$ is the number of videos.

\subsubsection*{Frame-level augmentation.}
Alongside with video-level features, the YouTube-8M dataset contains frame-level features: one preprocessed frame per second for every video. To use frame-level data in our models, we have tested several augmentation strategies. First, we computed and used as features various frame-level statistics, including the mean, standard deviation, median, minimum, maximum, mode, and video length (number of frames). These statistics were concatenated and used as input for our network architectures.

\begin{figure}[t]
\setlength\tabcolsep{2pt}
\begin{tabular}{ccc}
\includegraphics[width=.3\linewidth]{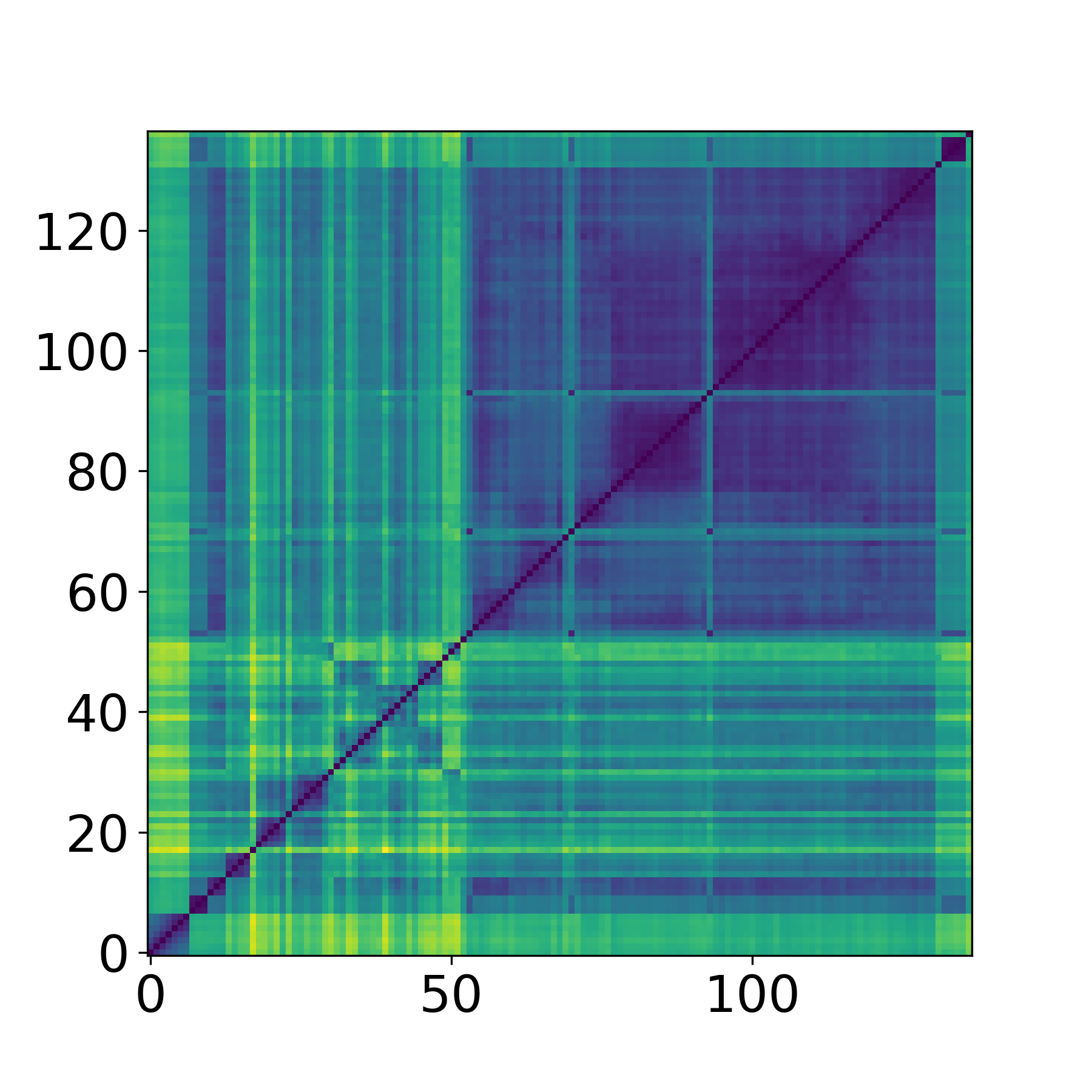} &
% верни как было) тут было как раз хорошо
\includegraphics[width=.3\linewidth]{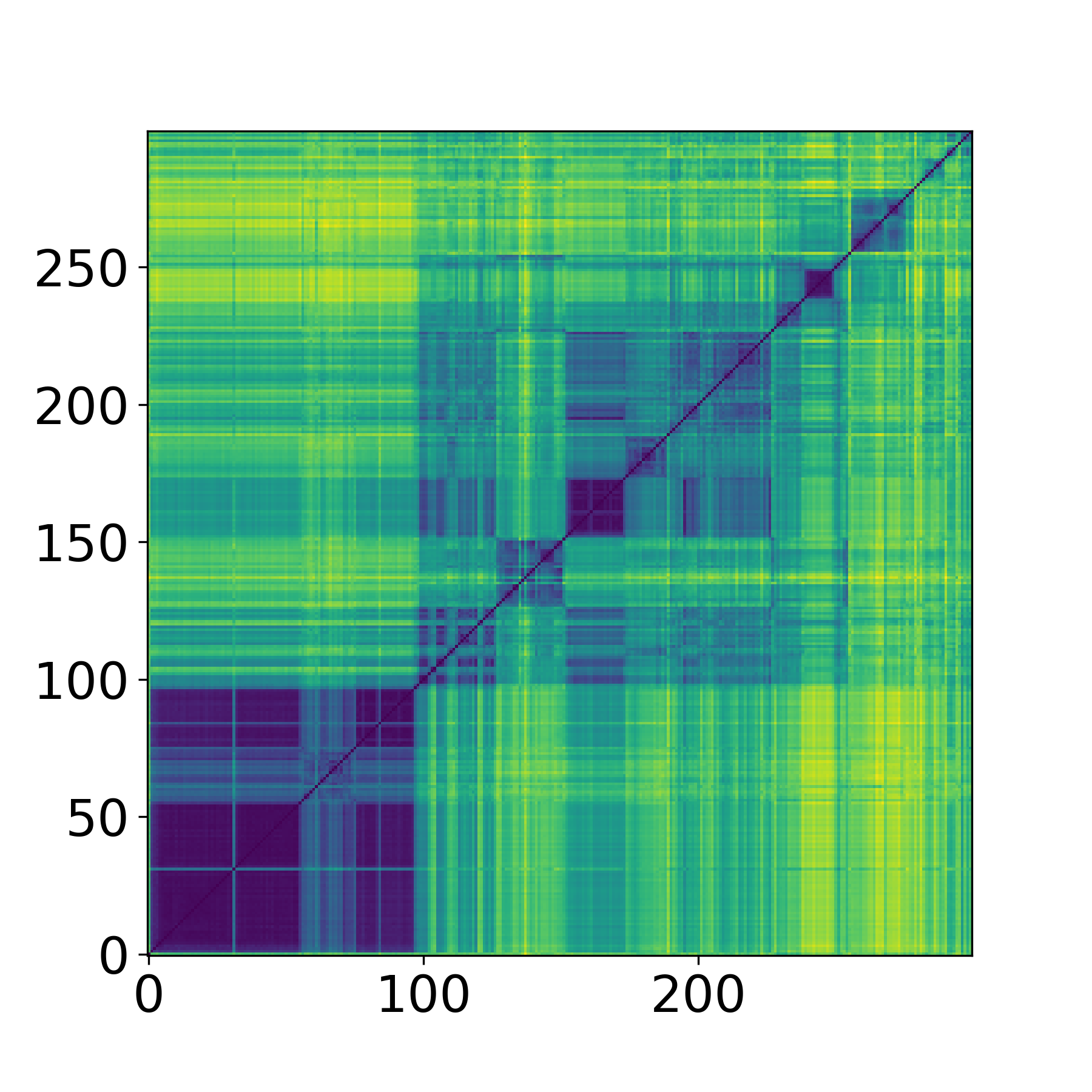} &
\includegraphics[width=.3\linewidth]{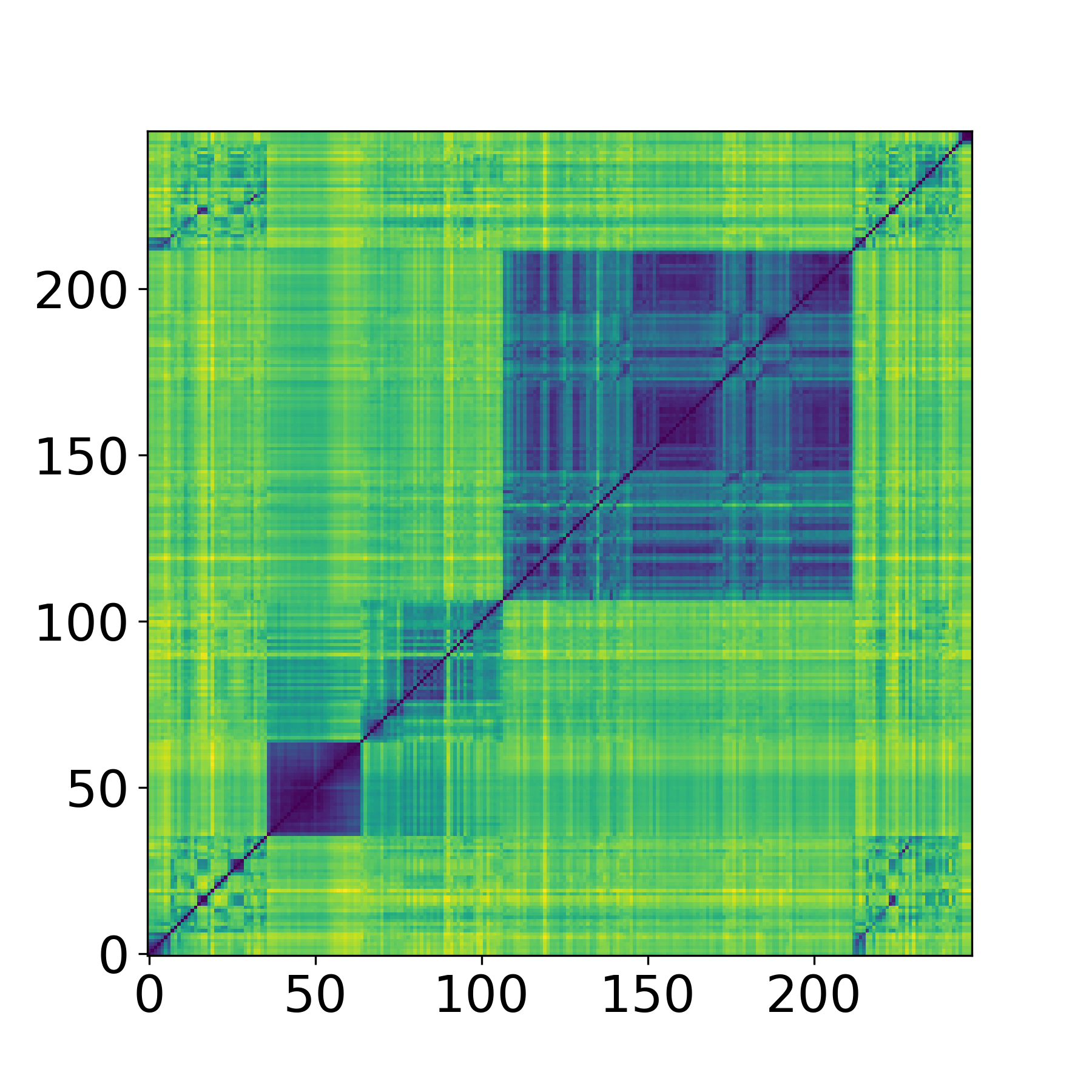}
\end{tabular}

\caption{Cosine distance between subsequent frames in three sample videos; we took abrupt changes in neighboring frames as evidence of a new ``scene'' beginning.}\label{fig:frames}
\end{figure}

To use frame-level features in the models, we computed the above statistics for different frames in the video. There are too many frames in a video, and neighboring frames are generally very similar, so it is detrimental to use all frame-level data and preferable to include only a few representative frames. We have tried several subsampling strategies to choose which frames to include as input:

\begin{enumerate}[(1)]
\item subsample a subsequence of frames at random or at regular intervals;
\item assuming that not all frames are equally important for training, we decided to take one frame per ``scene'' in the video; for this purpose, we cut the video into ``scenes'' by thresholding cosine distances between neighbouring frames (Fig.~\ref{fig:frames} shows sample correlations between the frames in a video; the scene structure is usually rather clear) and take one frame per scene;
\item apart from ``scenes'' specific for each video, we also found global centroids of frame-level features using large-scale $k$-means clustering for all frames in the dataset, and then used frame statistics from unique nearest centroids for the frames in a given video; with $10{,}000$ centroids, this approach yielded a $\gp$ score of $0.79$ (row~14 of Table~\ref{tbl:eval}).
\end{enumerate}

Another approach we tried for frame-level information was to try to capture just the \emph{dynamics} in a video sequence. The intuition behind this approach is that the main information about the video may be captured in what is changing between frames and not in the static component. To this end, we have tried several different techniques:
\begin{enumerate}[(1)]
	\item subtracting per-video average from every frame in the video;
    \item applying a high-pass filter using the \emph{db3} wavelet~\cite{Daubechies:1992:TLW};
    \item computing sparse component with Principal Components Pursuit~\cite{candes2011robust}.
\end{enumerate}
However, none of these techniques yielded a $\gp$ score greater than $0.5$.

Finally, the best technique that helped us incorporate frame-level information into video-level vectors in the best possible way is as follows. We have taken a number of linear combinations of frames with trainable coefficients (coefficients are linked to frame position) and then applied a ResNet-like model (see Section~\ref{sec:models}) to the concatenated vectors. This approach yielded a $\gp$ score of $0.855$ for $5$ linear combinations.

\input{models.tex}

\input{eval.tex}

\section{Conclusion}\label{sec:concl}

In this work, we have presented our approach for large-scale video understanding implemented as part of the 2nd YouTube-8M Video Understanding Challenge. Our general idea is to use a complex model, a large ensemble of heterogeneous neural networks, to produce soft labels for training set examples. The soft labels are then used as input for a much simpler model which is designed to fit into hardware requirements. We have shown the validity of our approach and presented a solution that took 4th place in the challenge.

The most promising direction for further work appears to be improving frame-level models: none of these models in our experiments outperformed video-level models (that have less information about the video), but they still provide important diversity for the ensemble, which shows that the information actually is there and can probably be extracted in a single first level model too. We also point out that the final model has very uneven quality across predicted labels; this probably means that in practice, different categories of labels may require different approaches.

\bibliographystyle{splncs}
\bibliography{egbib}
\end{document}

%% file: models.tex
\section{Models}\label{sec:models}

\subsubsection*{General approach.}
\begin{figure}[t]\centering
\includegraphics[width=\linewidth]{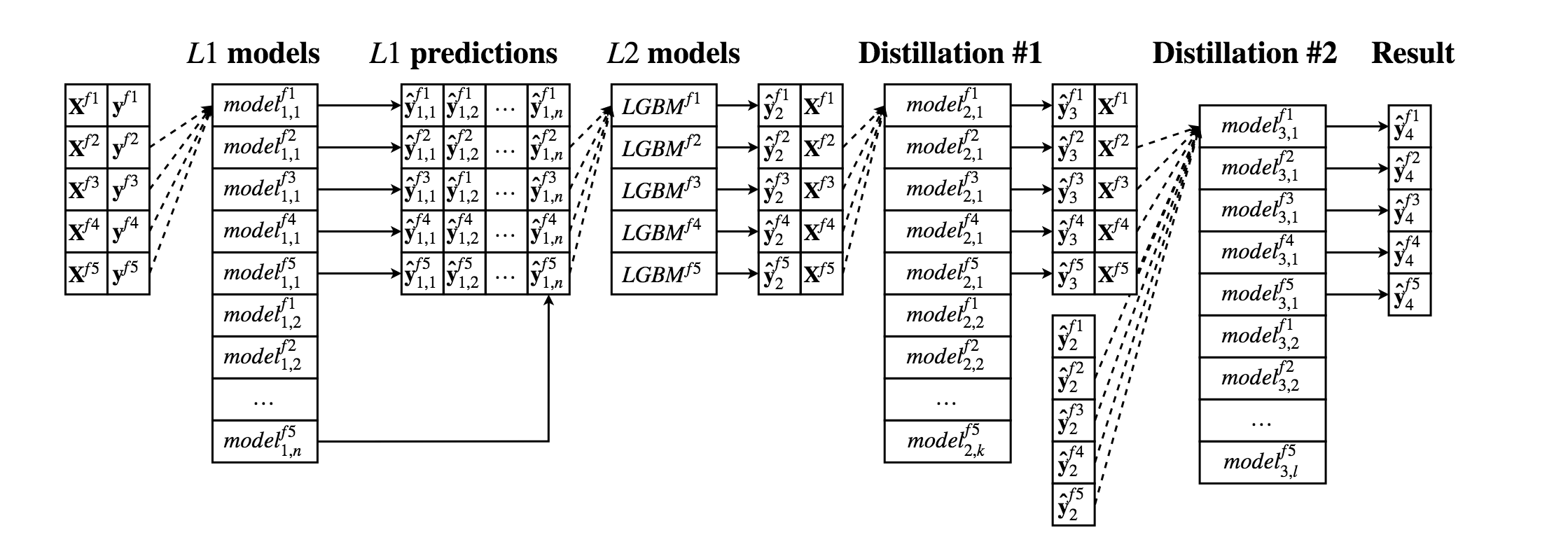}

\caption{General flowchart of our approach.}\label{fig:general}
\end{figure}

In this section, we define and discuss the models that we have tried in our experiments. In the YouTube-8M competition, one of the requirements for the final model was to fit it into 1Gb of memory. This is a rather limiting constraint for modern state of the art neural networks, so the objective was to both achieve the best possible results in terms of GAP and do it in limited space (and hence also limited inference time). To achieve both, we propose the following approach, illustrated on Figure~\ref{fig:general}.

First, we train various models on the video-level and frame-level features provided in the dataset, usually optimizing binary cross-entropy (BCE) with respect to the original hard labels; we call them \emph{first-level} models. The final first-level model is an ensemble based on different other first-level models.

Denoised \emph{soft labels} are then extracted from out-of-fold predictions of the ensemble: we take these predictions (in the form of $[0,1]$ confidence values) as soft labels. Then, we split the training set into five folds again and train simple fully-connected models with different depth, width, and activation functions.

Finally, we make out-of-fold predictions again, take each model's features on the penultimate layer (before final softmax), concatenate, and feed them into a new trainable classification layer. The feature extractors are frozen during training. This procedure yielded the final metric $\gp = 0.88729$.

Throughout this section, we refer to evaluation results summarized in Table~\ref{tbl:eval}, which shows whether the model uses frame-level features, its GAP@20 and BCE scores, and whether it became part of the final ensemble.

\begin{table}[t]\centering
\setlength{\tabcolsep}{5pt}
\begin{tabular}{|l|l|c|c|c|c|}\hline
 & \textbf{Model} & \textbf{Fr.} & \textbf{GAP@20} & \textbf{BCE} & \textbf{Ens.} \\\hline
 & \textbf{Final ensemble} & \checkmark & \textbf{0.88729} & --- & \checkmark\\
1 	& {ResNetLike  + soft labels} & $\times$ & 0.87417 & $9.2 \times 10^{-4}$ & \checkmark\\
2 	& {ResNetLike  + mixup} & $\times$ & 0.86105 & $9.7 \times 10^{-4} $  & \checkmark\\
3 	& ResNetLike over linear combinations & \checkmark & 0.85325 & $1.02 \times 10^{-3} $ & \checkmark\\
4 	& {ResNetLike + soft ranking loss} & $\times$ & 0.85184 & --- & \checkmark \\
5 	& AttentionNet & \checkmark & 0.85094 &  $1.08 \times 10^{-3} $&\checkmark\\
6 	& LSTM-Bi-Attention & \checkmark & 0.84645 & $1.04 \times 10^{-3} $& \checkmark\\
7 	& Time Distributed Convolutions & \checkmark & 0.84144 & $1.0 \times 10^{-3}$ & \checkmark\\
8 	& VLAD-BOW + learnable power & \checkmark & 0.83959 & $1.1 \times 10^{-3}$ & \checkmark\\
9 	& {Video only ResNetLike} & $\times$ & 0.83212 & $1.1 \times 10^{-3}$ & \checkmark \\
10 	& Time Distributed Dense Sorting & \checkmark & 0.83136 & --- & $ \times $ \\
11 	& EarlyConcatLSTM & \checkmark & 0.82998 & $1.2 \times 10^{-3}$ & \checkmark \\
12 	& Time Distributed Dense Max Pooling & \checkmark & 0.82656 & $1.1 \times 10^{-3}$ & \checkmark \\
13 	& Self-attention (transformer encoder) & \checkmark & 0.8237 & $1.2 \times 10^{-3}$ & \checkmark \\
14 	& 10000 clusters + ResNetLike & \checkmark & 0.7900 & $1.3 \times 10^{-3}$ & \checkmark \\
15 	& {Audio only ResNetLike} & $\times$ & 0.50676 & $2.5 \times 10^{-3}$ & \checkmark \\
16 	& {Bottleneck 4 neurons} & $\times$ & 0.41079 & $2.9 \times 10^{-3}$ & \checkmark \\ \hline
\end{tabular}
\caption{Evaluation results for basic models, ordered from best to worst.}\label{tbl:eval}
\end{table}

\subsubsection*{Video-level models.}
\begin{figure*}[p]
\setlength\tabcolsep{2pt}
\begin{tabular}{cc}
\begin{minipage}{.44\linewidth}
% \begin{figure}[t]\centering
\includegraphics[width=\linewidth]{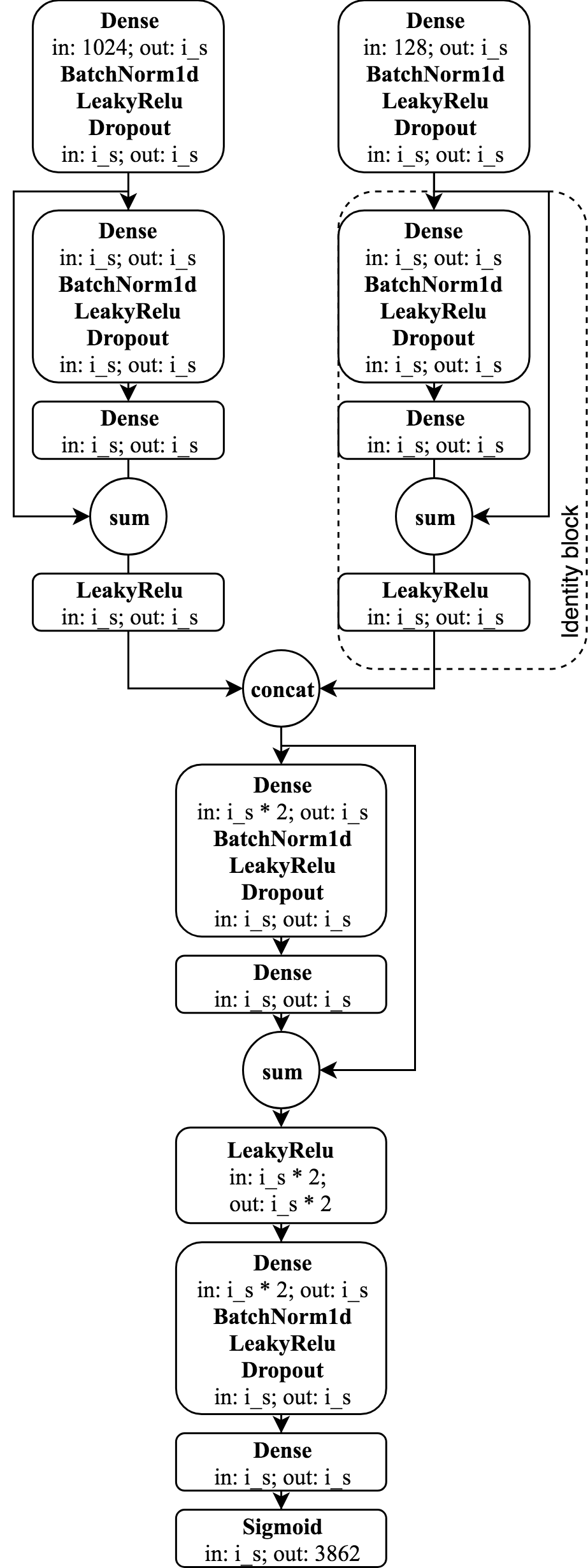}

\caption{ResNet-like architecture. i\_s, inner size, denotes the size of hidden layers.}\label{fig:resnet}
% \end{figure}
\end{minipage}
&
\begin{minipage}{.54\linewidth}
\begin{tabular}{c}
\includegraphics[width=.82\linewidth]{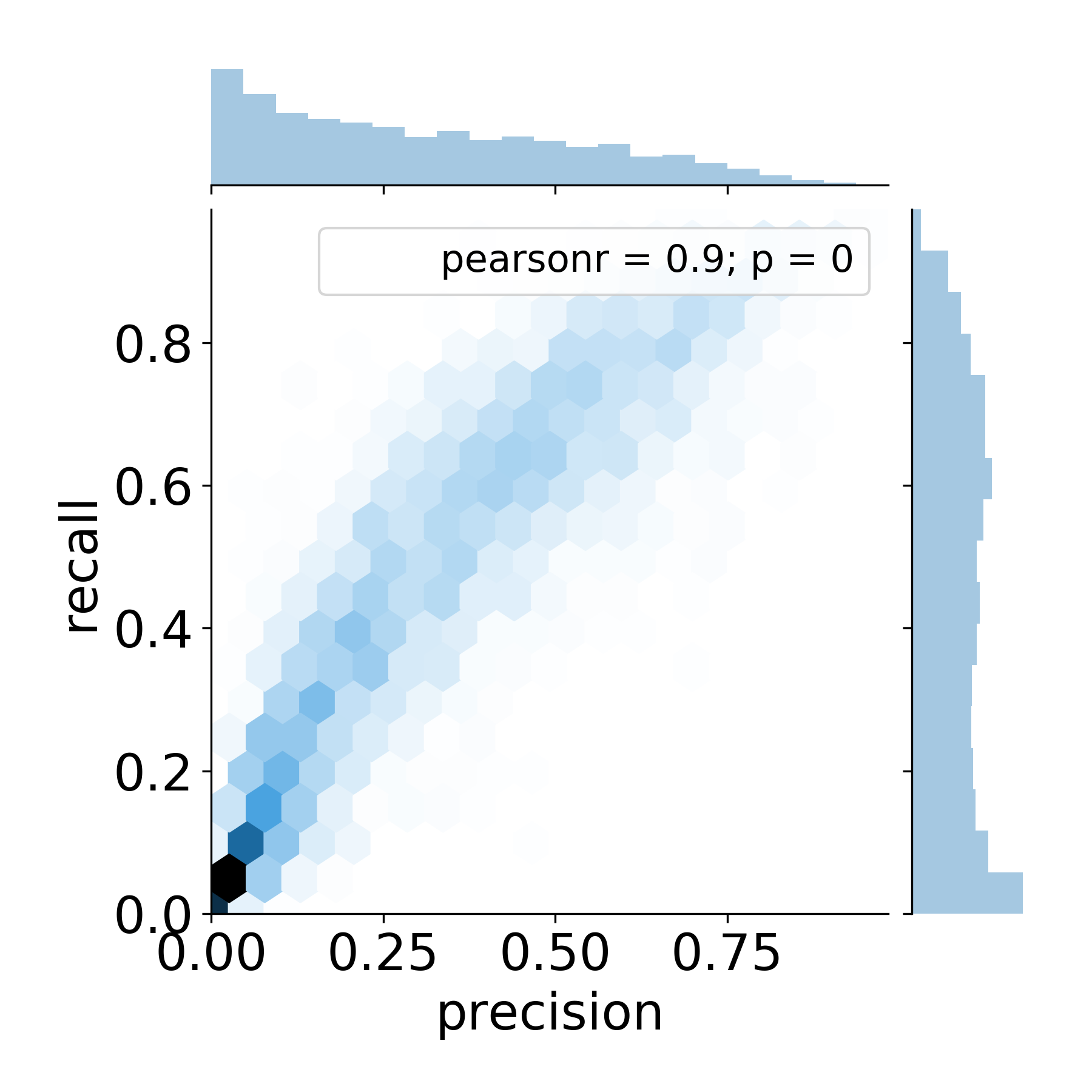} \\
(a) \\
\includegraphics[width=.82\linewidth]{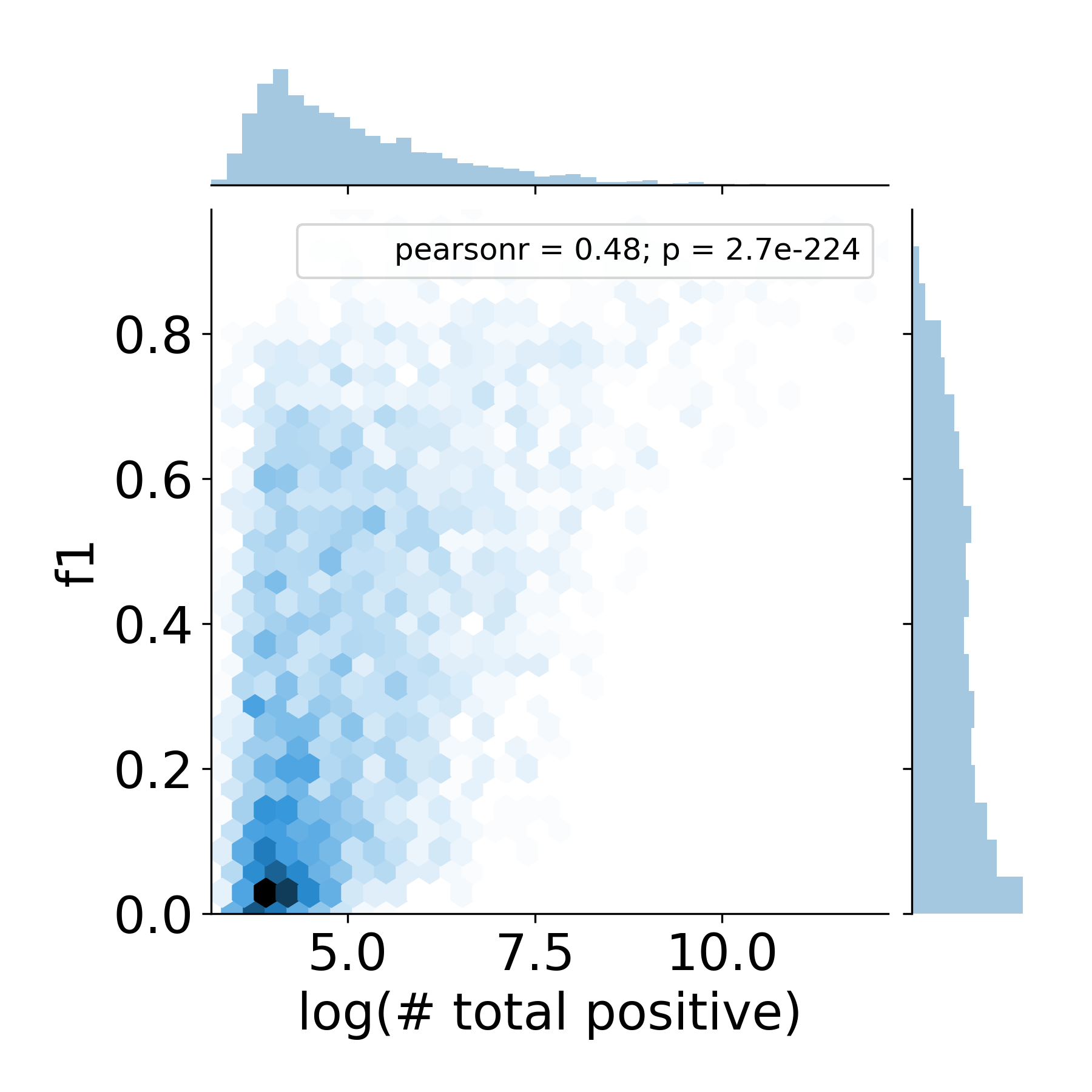} \\
(b) \\
\includegraphics[width=\linewidth]{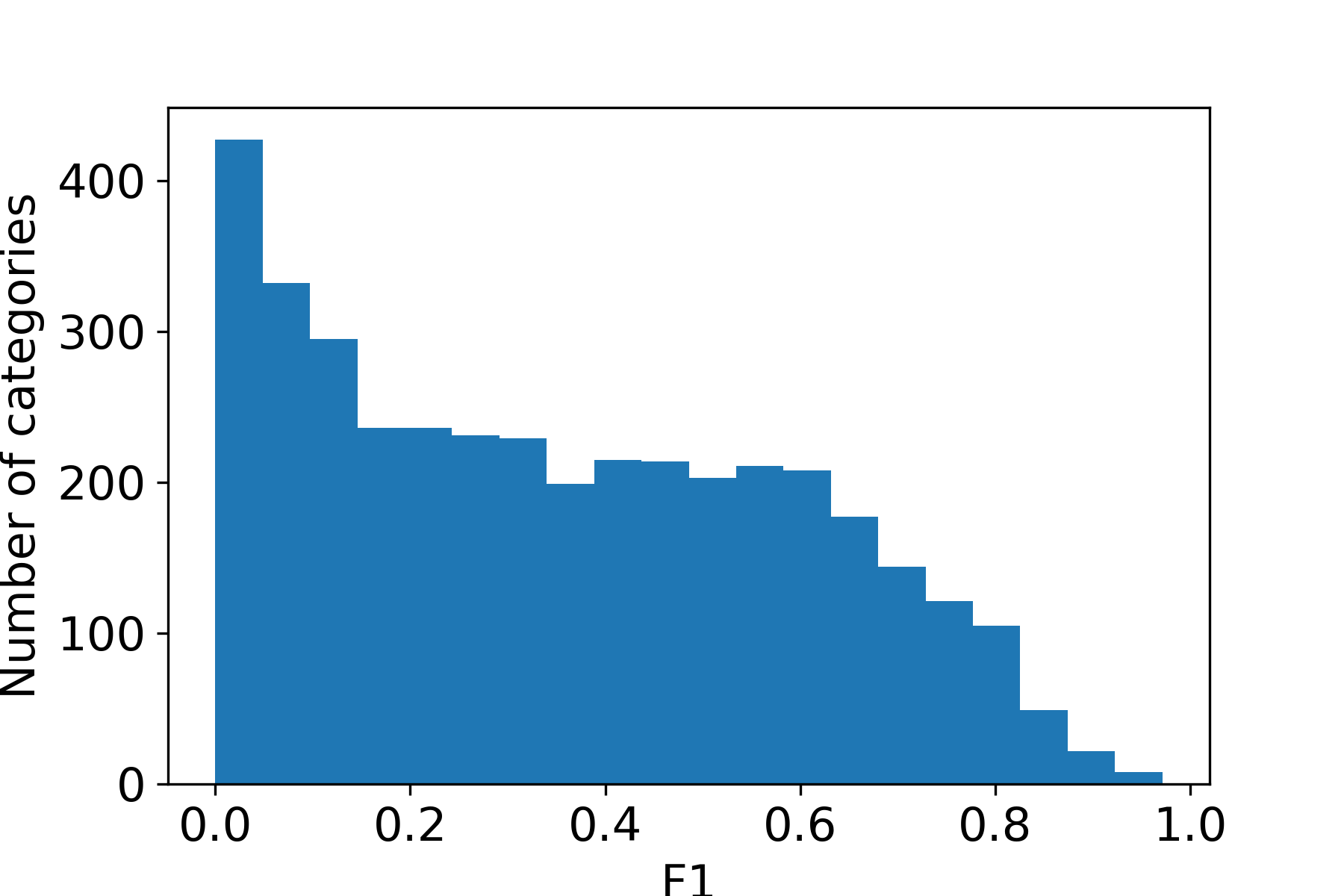} \\
(c)
\end{tabular}

\caption{Error analysis: (a) precision-recall heatmap; (b) F1-score as a function of the number of available training samples; (c) distribution of labels according to their final F1-score.}\label{fig:errors}
% \end{figure}
\end{minipage}
\end{tabular}
\end{figure*}

Our main first-level model is based on a ResNet-like architecture previously introduced for this problem in~\cite{n01z32017}; this architecture is shown on Fig.~\ref{fig:resnet}. The model takes three hyperparameters as input: inner\_size, av\_id\_block\_num, concat\_id\_block\_num, and dropout.
By default we used inner\_size=2048, av\_id\_block\_num=1, concat\_id\_block\_num=1, dropout=0.5. 

We have not found any strong correlation between network depth and final GAP value during our tests, yet our best first-level validation score $\gp=0.86105$ was achieved by a ResNet-like model with parameters $(4, 4, 0.4)$ and mixup $\alpha=0.3$, trained with binary cross-entropy as the loss function (row~3 of Table~\ref{tbl:eval}).

\subsubsection*{Mixup}
One important idea that helped improve first level models was \emph{mixup}, a recently developed new regularization technique~\cite{zhang2018mixup}. Mixup has been empirically shown to deliver better accuracy on the validation set than original labels. The mixup method produces ``virtual'' training samples as linear combinations of existing training samples and their targets: 
$$x = \lambda x_i+(1 - \lambda)x_j,\quad y = \lambda y_i+(1 - \lambda)y_j,$$ where $(x_i, y_i)$ and $(x_j, y_j)$ are feature-target vectors sampled from training data, and $\lambda\in[0,1]$ is drawn from a beta distrubution, $\lambda\sim\mathrm{Beta}(\alpha,\alpha)$, where $\alpha$ is the main parameter of mixup; in our experiments we set the mixup $\alpha$ parameter to $0.4$. Most of our first level models use mixup, and it improves results significantly; e.g., the basic ResNetLike architecture with mixup has $\gp=0.86105$ (row~2 of Table~\ref{tbl:eval}) compared to $\gp=0.85846$ without it.

\subsubsection*{Recurrent frame-level models.}
Along with aggregated video-level features, the dataset also provides temporal frame-level representation of the videos. 
% After analyzing last year solutions [LINKS TO WORKSHOP PAPAERS] we came up with a number of different approaches. Our team conducted 
We have conducted multiple experiments with recurrent neural networks based on LSTM and GRU units to incorporate temporal features into the model. We have tried both unidirectional and bidirectional LSTM-based networks with 2 hidden layers size of 1024 followed by fully-connected layer size of 2048; we call this basic RNN model \emph{EarlyConcatLSTM} because it receives as input concatenated audio and video features (row~11 of Table~\ref{tbl:eval}).

Another approach was to use a learnable bag-of-words representation. We used the \emph{VLADBoW} model proposed in~\cite{miech2017learnable}. One major difference was to introduce a learnable power coefficient, that is, we learned it as
$$
\text{BOW}(k) = \sum_{i=1}^{N}a_k(y_i)\text{~~with~~}
y_i=[Wx_i+b]_+^p\text{~~and~~}
a_k = \frac{e^{y_{ik}}}{\sum_{j=1}^{k}e^{y_{ij}}}
$$
with trainable $p$, getting as a result $p=0.628$. This model achieved $\gp=0.83959$ (row~8 of Table~\ref{tbl:eval}).

\subsubsection*{Attention-based frame-level models.}
Recent research indicates that attention-based networks often outperform classical RNNs in similar tasks~\cite{NIPS2017_7181}. We have tried several attention-based architectures:
\begin{enumerate}[(1)]
\item encoder architectures similar to the ones from the \emph{Transformer} model~\cite{vaswani2017attention} yielded $\gp = 0.8237$ (row~13 of Table~\ref{tbl:eval}); we used stacked multihead self-attention mechanism with and without position coding, and Transformer without position coding was much more successful; this may indicate that most labels could be detected by taking into account individual frames and disregarding relations between frames and their mutual temporal positions;
\item \emph{stacked attention} model, which contains multiple layers, each computing a single attention vector for a sequence, concatenates them to obtain the feature vector at every time step, and then performs global average pooling, reaching a very high $\gp = 0.85094$ (row~5 of Table~\ref{tbl:eval}).
\end{enumerate}

\subsubsection*{Time-distributed models.}
One more group of approaches is to use time-distributed dense layers with different poolings to finally convert them to a single vector; we have used three different approaches with time-distributed layers. Our simplest model uses a single fully connected layer and frame-wise max pooling afterwards. This approach yields $\gp=0.82656$ (row~12 of Table~\ref{tbl:eval}). 
The second approach is to apply a ResNet-like model to every frame in a video with additional two dense layers on the cumulative output of every per-frame model sorted by confidence; this gave us $\gp=0.83136$ (row~10 of Table~\ref{tbl:eval}).
Finally, we tried time-distributed convolutional layers; the best convolutional model contains several layers of convolutions followed by max-pooling for video and audio separately, then concatenating the resulting features. This approach yields $\gp=0.84144$ (row~7 of Table~\ref{tbl:eval}).

\subsubsection*{Loss function selection.}
As the main loss function we used the binary cross-entropy (BCE). However, since the primary objective was in terms of a ranking metric, some models were trained using the soft ranking loss; in our experiments, for equivalent models this loss has led to slightly worse GAP scores than BCE. Reweighting BCE using weights derived from mispredicted examples did not show any significant improvements in validation score.

The global average precision metric is based on the ranking among predictions rather than their specific scores. Thus, we applied batch-wise ranking in the following way. First we extract top 30 scores corresponding to negative labels for every sample in the batch. Then we extract all scores corresponding to positive labels in the batch. Finally, we apply the following pairwise ranking loss:
$L(p_i,n_j) = \log(1 + \exp(n_j-p_i+1))$,
averaging it across all pairs of scores in the batch. This approach led to, e.g., $\gp=0.85184$ (row~4 in Table~\ref{tbl:eval}) for ResNetLike compared to $\gp=0.85325$ without the ranking loss (row~3), but ranking-based models were still useful for the final ensemble.

We also experimented with standard hinge ranking loss function, but found that hard thresholding is harmful for model convergence and the GAP metric. We have also tried the following related ideas with no strong positive effect (we believe that in this case it is important to report negative results as well):
\begin{enumerate}[(1)]
\item penalizing predicted confidence for a label by cross-validation classification accuracy for that label (the smaller the accuracy, the stronger the penalty); the intuition here is that noisier labels must receive marginally less confident scores;
\item inducing noise by flipping labels at random (with higher probability for labels with smaller cross-validation accuracy);
\item loss function correction as proposed in~\cite{NIPS2013_5073};
\item dropout technique from~\cite{DBLP:journals/corr/JindalNC17};
\item test time dropout with averaging the predictions.
\end{enumerate}

\subsubsection*{Final ensemble and distillation.}
Our general approach, as shown in the beginning of this section, implies a general ensemble model which is then used to obtain soft labels (similar to the knowledge distillation approach as shown in~\cite{hinton2015distilling}) used instead of noisy ones to improve the final model's quality.

We used an ensemble of 115 first level models to prepare the final model. On the first level, we used 95 video-level and 20 frame-level models; all of them are neural networks due to the origin and nature of the data provided. However, the ensemble itself was done with a gradient boosting model, namely LightGBM~\cite{NIPS2017_6907}. Distilled soft labels produced by this ensemble allowed us to achieve $\gp = 0.88729$ with the final compressed model.

%% file: eval.tex
\section{Experimental Evaluation}\label{sec:eval}

In this section we present our experimental results on the YouTube-8M dataset.

\subsubsection*{Data preparation and augmentation.}
To make the resulting models more robust, we initially shuffled all data and split it into $5$ cross-validation folds to train and validate all our models. Throughout the paper, we estimate the validation score as GAP measured on the hold-out fold, and the final GAP scores shown in Table~\ref{tbl:eval} are calculated as an average of $5$ folds.

The effectiveness of data augmentation in various tasks, especially related to computer vision, has been known and empirically validated for a long time~\cite{DBLP:journals/corr/abs-1003-0358,DBLP:journals/corr/abs-1712-04621,DBLP:journals/corr/abs-1708-06020}. Despite the fact that the initially provided features were preprocessed embeddings rather than images of frames, we performed several experiments with data augmentation techniques. In particular, we tested statistical bootstrap, using a weighted average for a random subset of frames and mixup~\cite{zhang2018mixup}. Some models were trained on random subsets of available features. 

\subsubsection*{Compact representations and single modality experiments.}
We have conducted experiments to see how performance (in terms of the validation metric) of the model degrades depending on the size of the inner representation within our model. Using only $4$ neurons in hidden layers has allowed us to achieve GAP equal to $0.41079$.

We have also trained separate audio-only and video-only video-level models based on a single modality only, achieving the best $\gp$ scores of $0.83493$ and $0.50676$ respectively. These models were included into the final ensemble of first-level models for the sake of diversity.

\subsubsection*{Training parameters.}
All models in our comparison were trained and validated with multiple NVIDIA P40 GPUs. In our experiments, we have attempted to tune the mini-batch size and learning rate. The experiments have shown the following tradeoff: lower mini-batch size increases the convergence rate but the final quality of the model can deteriorate slightly. At the same time, lower batch sizes mean that an experiment takes up significantly less GPU memory (the memory footprint is basically linear in the batch size), which is an important argument for lower batch sizes in practice. Note that larger mini-batch sizes require warmup which does not let one use high learning rates right away~\cite{DBLP:journals/corr/GoyalDGNWKTJH17}.

\subsubsection*{Results.}
The main results of evaluating our models are summarized in Table~\ref{tbl:eval}. We have already discussed our models in detail in Section~\ref{sec:models}. One general result that can be seen from Table~\ref{tbl:eval} is that, rather surprisingly, video-level models on the first level consistently outperform frame-level models despite the fact that they receive far less information about a given video. This probably means that there is still a lot of work to be done to improve recurrent and/or attention-based approaches to video understanding---after all, our final ensemble does use frame-level models and hence this information. Moreover, frame-level models have high importance in the gradient boosting model, which shows that they add a lot of diversity and new information to the ensemble.

\subsubsection*{Error analysis}
To analyze the quality and characteristic features of our final model, we performed error analysis according to the following methodology. We take top $20$ predicted labels for a video and classify these labels into true positive, false positive, and false negative classes. A label is true positive if it is positive and it has predicted score higher than all negative labels for that video; a label is false positive if it is negative and has predicted score higher than at least one positive label for that video; finally, a label is false negative if it is positive and has predicted score less than at least one negative label for that video.

\begin{figure}[t]\centering
% \setlength\tabcolsep{2pt}
% \begin{tabular}{cc}
% \includegraphics[width=.4\linewidth]{images/5_precision_recall_2d_histogram.png} &
% \includegraphics[width=.4\linewidth]{images/6_f1_logpositive_2d_histogram.png} \\
% (a) & (b) \\
% \includegraphics[width=.45\linewidth]{images/4_f1_category_histogram.png} &
\includegraphics[width=.85\linewidth]{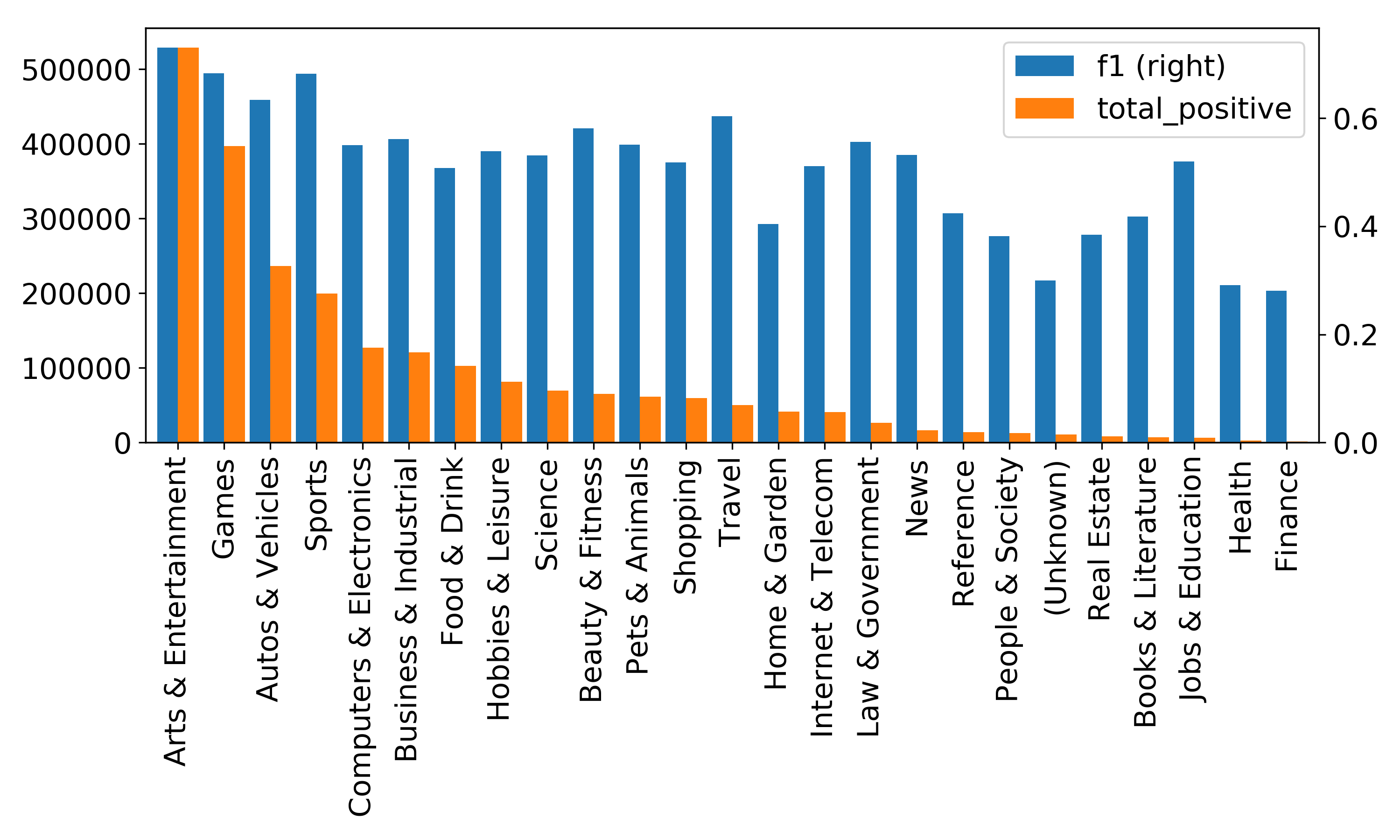}
% (c) & (d) \\
% \end{tabular}

\caption{Classification accuracy across verticals.}\label{fig:acc}
\end{figure}

Figure~\ref{fig:errors} shows the error analysis for our ResNet-like model distilled against the ensemble. Fig.~\ref{fig:errors}a shows that false positives are marginally more probable than false negatives, but not by much, which shows that the model is well balanced. Fig.~\ref{fig:errors}b illustrated an expected effect: the more training examples we have, the higher is the accuracy; note, however, that the final model can also accurately predict many rare labels. Nevertheless, Fig.~\ref{fig:errors}c shows that vice versa, there are many categories that the model cannot predict accurately; these are mostly rare categories, but for some the reasons might be different and might require further study. Finally, Fig.~\ref{fig:acc} shows classification accuracy across verticals (coarse-grained labels) in comparison to the number of positive examples in the corresponding training subset (one fold). Again, we see that generally more popular categories are easier to predict, but the effect varies significantly.